# Creating and managing large annotated parallel corpora of Indian languages

Ritesh Kumar[1], Shiv Bhusan Kaushik[2], Pinkey Nainwani[1], Girish Nath Jha[2]

[1]Centre for Linguistics, [2]Special Centre for Sanskrit Studies

Jawaharlal Nehru University, New Delhi

E-mail: {riteshkrjnu, shivkaushik.engg, pinkeybhu39, girishjha}@gmail.com

**Abstract**

This paper presents the challenges in creating and managing large parallel corpora of 12 major Indian languages (which is soon to be extended to 23 languages) as part of a major consortium project funded by the Department of Information Technology (DIT), Govt. of India, and running parallel in 10 different universities of India. In order to efficiently manage the process of creation and dissemination of these huge corpora, the web-based (with a reduced stand-alone version also) annotation tool ILCIANN (Indian Languages Corpora Initiative Annotation Tool) has been developed. It was primarily developed for the POS annotation as well as the management of the corpus annotation by people with differing amount of competence and at locations physically situated far apart. In order to maintain consistency and standards in the creation of the corpora, it was necessary that everyone works on a common platform which was provided by this tool.

## 1. Introduction

In recent time, availability of huge annotated corpora has become very essential for the development of language technologies for any language. Since almost all the Indian languages are considered less-resourced languages, it is very necessary to develop extensive language resources for them in order to make them technologically strong and efficient.

This paper presents the challenges in creating and managing large parallel corpora of 12 major Indian languages (which is soon to be extended to 23 languages) as part of a major consortium project funded by the Department of Information Technology (DIT), Govt. of India (Jha, 2009 and Jha, 2010), running parallel in 10 different universities of India (Choudhary & Jha, 2011).

In order to efficiently manage the process of creation and dissemination of these huge corpora, the web-based (with a reduced stand-alone version also) annotation tool ILCIANN (Indian Languages Corpora Initiative Annotation Tool) has been developed. It was primarily developed for the POS annotation as well as the management of the corpus annotation by people with differing amount of competence at locations physically situated far apart. In order to maintain consistency and standards in the creation of the corpora (it is essential for any corpora to be usable in NLP), it was necessary that everyone works on a common platform which was provided by this tool. The use of the tool ensured that the data is saved on a centralized server in a uniform format which could be later utilized for any NLP task. Besides providing administrative and annotation facilities, the tool also provides the facility for creating parallel corpora as well as automatically adapting the linguistic data from any other source. It could also be potentially used for crowd-sourcing the annotation task and creation of language resources for use in NLP.

## 2. The Corpora

In its first phase, Indian Languages Corpora Initiative (ILCI) was involved in the creation of translated parallel corpora of 12 Indian languages viz., Hindi, Bangla, Oriya, Urdu, Punjabi, Marathi, Gujarati, Konkani, Telugu, Tamil, Malayalam and English. The basic data of the corpora was collected in Hindi (sourced from different written texts like magazines, newspapers, books, etc.) and then it was manually translated into all other languages by the respective language experts. The data was collected from the two domains health and tourism with 25,000 sentences in each domain (with an average length of 16 words per sentence, counting up to around 400,000 words in each domain in each language). The sentences in each language are aligned parallel (along with an alignment up to the word level as far as possible) and each sentence is given a unique ID (details of data collection and the corpora are included in Choudhary and Jha (2011)). In the second phase of the project, 50,000 sentences are being added to the corpus of each language from the two domains agriculture and entertainment.

The corpus of every language has been initially annotated with part of speech information. This work was done manually by each language group. However soon the challenges of annotating such huge corpora by the people physically distributed over different areas began to come to the fore. Two of the major challenges included:

1. It was very difficult to maintain the sanity and uniformity of the data across all the groups since the annotation was being carried out by the people of varying degree of experience and expertise. In such a scenario the annotated data did not carry a uniform structure despite the clear instructions on how to carry out the annotation. Since it was very necessary to maintain the uniformity throughout the corpora so that any meaningful work could be done using these corpora, there was an urgent need to devise a mechanism to ensure this.

2. There also were some very administrative issues that needed some urgent attention. These included keeping track of the progress of every language group and ensuring that the work is completed within the stipulated time period by each member of the consortium.

As a result of these challenges, the idea of a web-based application for managing as well as carrying out the annotation task came to the fore.

### 3. Managing the Parallel Corpora

Over the last two decades, numerous annotation tools have been created to meet the required demands of various projects. The most popular and well-known annotation tool among them is General Architecture for Text Engineering (GATE) (Cunningham, 2011). However since it is a stand-alone application, it does not provide the facility of managing a physically distributed project. Moreover it also does not have the facility of managing and creating a parallel corpus. Some of the other significant tools include Stuhrenber et al., (2007), Russell et al. (2005), besides numerous others. Bird et al., (2002) came up with a tool which deals with annotations called ATLAS (Flexible and Extensible Architecture for Linguistic Annotations). Kaplan et al. (2010) discussed SLATE (Segment and Link-based Annotation Tool Enhanced) in their paper. It is a web-based annotation tool and addresses 10 annotation needs: (1) managing the role of annotator and administrator, (2) delegation and monitoring work, (3) adaptability to new annotation tasks, (4) adaptability within the current annotation task, (5) diffing and merging (diffing and merging of data from multiple annotators on a single resource to create a gold standard), (6) versioning of corpora, (7) extensibility in terms of layering, (8) extensibility in terms of tools, (9) extensibility in terms of importing/exporting and, (10) support for multiple languages.

However none of these tools are meant to support the requirements of creating and managing translated parallel corpora. Besides the annotation needs mentioned by Kaplan et al. (2010), couple other requirements need to be fulfilled by a tool for creating and managing parallel corpora include -

1) Translation Work: To build parallel annotated corpora, the tool should support the translation of the source data in the respective languages.

2) Quality assurance: It is a key concern as far as full crowd sourcing is concerned but there must be some automated features to check the quality of translated and tagged data.

3) Crowd-sourcing: The tool must be flexible enough to adapt to the needs of crowd-sourcing at any given point of time.

### 4. The ILCI Annotation Tool

ILCIANN is a server-based web application which could be used for any kind of word-level annotation task in any language. It is developed using Java/JSP as the programming language and is running on Apache Tomcat 4.0 web server. Some of the facilities provided by the tool for managing a large project include the following:

1. User Management and Monitoring Facility: The tool recognises users at three hierarchical levels:

a. Master Admin: The master admin is basically the main administrator of the project, who spearheads the project and overlooks all the language groups working in the project. The major responsibilities of master admin include

i. Uploading the Files: The major responsibility of the master admin is to upload the source files which are to be translated in different target languages. (S)he is also required to upload the translated files in different languages for annotation (if the translation is not done in the tool).

ii. User Management: This step involves creating the login of users who would annotate the data. Only the master administrator has the authority to create/delete/modify the login for the users who are supposed to annotate/translate the data as well as for the individual language administrators. It ensures the safety as well as authenticity of the tagged data, while theoretically giving an opportunity to a huge community to support and help in building language resources for their languages. Further, since the project of creating parallel corpora, by definition, involves multiple languages, therefore, the users and administrators are also assigned to the language on which (s)he is supposed to work. For instance, if x is a Hindi language annotator, (s)he can only work on Hindi data and cannot do any modification (tagging the data, editing the data and saving it) in other language files.

iii. Monitoring the Project: Besides this, the master admin can maintain the time log of the user accounts (which include the details about currently logged in users, login and logout history of different users from any language group), monitor the overall progress of the project (including the amount of work completed), send notices and reminders to the users as well as administrators of individual language groups regarding the progress of the project.

iv. The other functions and facilities of the master admin are the same as for the administrator of individual language groups, discussed below.

b. Administrator (Admin) – For the purpose of management, each language group is assigned to an administrator The following responsibilities are given to an admin in order to facilitate the increased productivity and proper administration of the project:

i. Assigning and Monitoring the Work: The admin could assign a set of maximum 3 files for annotation to a single user at one time (and a new file is assigned only after one of the files is completed). It eliminates any scope for the duplication of effort in a huge project and also ensures that one or more files are not left incomplete. Furthermore it also helps in quality control of the annotation work by ensuring that, in general, only one user works on one file (and even if a file is re-assigned to some other user then a record is maintained). It also helps in keeping a record of the progress as well as the precise achievement of the individual annotators/translators in the project.

ii. Downloading the Files: The files could be downloaded only when each sentence of the file is tagged and only the administrator has the right to download the files.

c. User (Annotator/Translator) – The user is responsible only for the annotation/translation task. They can work on one of the files that have been assigned to them and annotate/translate it. The users, along with the administrators are able to view the progress made in the project in terms of total work that needs to be done and the total work that has been currently completed.

2. Annotation Facility: In its current form, the tool allows the user to annotate the data at the word level. Some of the major features of the annotation facility include:

a. Complete Language and Tagset Independence: There is no restriction at all related to the use of tagset or language and in any given project any tagset could be used for annotation.

b. Limited Intelligence: The tool provides the facility of limited automatic tagging for closed grammatical categories like pronouns, postpositions, conjunctions and quantifiers which reduces the burden of human annotators. The list of words marked for automatic tagging could be modified and edited by the user during the annotation process and the changes made by one user become available to all other users working in the same language in real time.

c. Limited Editing: The users are also allowed to edit the

data in case they find some errors (related to the structure, orthography, translation, etc) or they see a mismatch with the source data.

d. Quality Control: In order to ensure that the data is saved properly and the annotation is carried out using only valid tags from the available tagset, the users are presented with an option to choose one of the tags from the tagset and are not given any freedom in assigning the tags (this will prove to be inefficient for very large tagsets and the effort is made to improve it).

3. Translation Facility: The work is under progress to include the facility of translation of the source data into several target languages in the tool. In its initial stage this facility is expected to provide a rough translation with the help of a bilingual dictionary to help the translators and increase their productivity.

4. Adaptation Facility: The tool also has the facility to adapt and modify data from other sources as well as noisy data in such a way that it could be used properly for the annotation work.

## 5. Conclusions and Future Work

The use of ILCIANN for annotation purposes could help in resolving lots of issues both on the side of the annotators as well as the developers as we could see in the case of the ILCI project. On the one hand it ensures the uniformity of the data without any scope for any noise creeping into it, which becomes inevitable if the annotation work is carried out manually by a large number of annotators. This makes things easier for the developers who want to work with the data. At the same time, since the tool is a web application, a huge number of people could work together in parallel and seamlessly (without actually worrying about what others have completed, since the tool by its very structure eliminates any scope of redundancy) and contribute to the development of the language resources. Thus it could prove to be a very significant tool for creating annotated corpora, especially in smaller and less-resourced languages, with the help of the community and a large number of online contributors.

The tool needs to be further developed to ensure automated checks for quality assurance (always a concern with online crowd sourcing), check the inter-annotator agreement, increase the options for importing/exporting the data in different formats and also include the facility to create corpora from the web automatically.

## 6. Acknowledgement

The project for creating the parallel corpora as well as the annotation has been funded by the Department of Information Technology (DIT), Govt. of India. We are thankful to them for their co-operation and support. We are very thankful to all the members of the project for their immense contribution to the project . We are also very thankful to the reviewers and the proofreaders of this paper for their invaluable suggestions and improvements.